\documentclass{article}

\usepackage{times}  
\usepackage{helvet}  
\usepackage{courier}  
\usepackage{url}  
\usepackage{graphicx}  
\usepackage{mathtools}
\usepackage{booktabs}
\usepackage{amsmath,amssymb,amsfonts}

\usepackage{caption}
\usepackage{graphicx}
\usepackage{subfigure}
\usepackage{amsthm,dsfont}
\usepackage{algorithmic}
\usepackage[ruled, noend]{algorithm2e}
\newcommand{\ignore}[1]{}
\usepackage{color}
\usepackage{authblk}
\usepackage{graphicx}
\usepackage{float} 
\usepackage{multirow}
\usepackage{amsfonts}
\usepackage{comment}
\usepackage{fancyhdr}
\usepackage{rotating}
\newtheorem{theorem}             {Theorem}

\begin{document}
	
	\title{Heuristic Strategies for Solving Complex Interacting Stockpile Blending Problem with Chance Constraints}

\author{Yue Xie}
\author{Aneta Neumann}
\author{Frank Neumann}
\affil{Optimisation and Logistics, School of Computer Science,\\ The University of Adelaide, Adelaide, Australia}
\renewcommand\Authands{ and }
\maketitle
\begin{abstract}
Heuristic algorithms have shown a good ability to solve a variety of optimization problems. Stockpile blending problem as an important component of the mine scheduling problem is an optimization problem with continuous search space containing uncertainty in the geologic input data. The objective of the optimization process is to maximize the total volume of materials of the operation and subject to resource capacities, chemical processes, and customer requirements. In this paper, we consider the uncertainty in material grades and introduce chance constraints that are used to ensure the constraints with high confidence. To address the stockpile blending problem with chance constraints, we propose a differential evolution algorithm combining two repair operators that are used to tackle the two complex constraints. In the experiment section, we compare the performance of the approach with the deterministic model and stochastic models by considering different chance constraints and evaluate the effectiveness of different chance constraints.

\end{abstract}

\section{Introduction}
Mining is the extraction of economically valuable minerals or materials from the earth. This has raised the importance of the production scheduling process due to its significant role in the profitability and efficiency of any mining operation. Mine production scheduling problem \cite{moreno2017linear} is a well-study mining engineering problem, and it has received much attention in past decades \cite{osanloo2008long} from both engineering and research. The task of the mine production schedule is to generate a mining sequence and ensure the product meets the blending resource constraints and object to maximize the net present value of the mining operation.

The mine production scheduling problem is commonly formulated as a Mixed-Integer Program (MIP) with binary variables \cite{johnson1968optimum,bley2012solving,Topal}. However, since it becomes a challenge for the MIP when the problem deals with the blending resource constraints. Lipovetzky et al.~\cite{Nir2014} introduced a combined MIP for a mine planning problem, which devises a heuristic objective function in the MIP and can improve the resulting search space for the planner. Samavati et al. \cite{samavati2017local} proposed a heuristic approach that combines local branching with a new adaptive branching scheme to tackle the production scheduling problem in open-pit mining.

Stockpiles are essential components in the supply chain of the mining industry, and play a significant role in the mine scheduling problem. Jupp et al. \cite{jupp2013role} introduced the four different reasons for stockpiling before material processing: buffering, blending, storing, and grade separation. In open-pit mine production scheduling problem, stockpiles are used for blending different grades of material from the mine or keeping low-grade ore for possible future processing \cite{moreno2017linear,rezakhah2020practical}. Rezakhah et al. \cite{rezakhah2019open} used a linear-integer model to approximate the open-pit mine production scheduling with stockpiling problem which forces the stockpile to have an average grade above a specific limit. Recently, some researchers present nonlinear-integer models to solve open pit mine production scheduling with stockpiles. Tabesh et al. \cite{tabesh2015comprehensive} proposed a nonlinear model of stockpiles to optimize a comprehensive open-pit mine plan but not give any results. Bley et al. \cite{bley2012solving} proposed a nonlinear model for mine production planning, however, they only consider one stockpile.  

In this paper, we study an important component of the mine production scheduling problem, the stockpile blending problem. This problem is challenging to address in terms of blending material from stockpiles for parcels to match the demands of downstream customers. We define the stockpile blending problem as an optimization problem that aims to maximize the volume of valuable material from all parcels by finding the percentage that each stockpile provides for each parcel in the whole planning. Furthermore, the strategy has to respond to the mining schedule and the market plan where the mine schedule provides the material mining and sending to corresponding stockpiles in each period, and the market plan provides the customer requirements. 

Solving the stockpile blending problem in mining optimally is critical because it is based on an uncertain supply of mineralized materials for the resource available in the mine. This uncertainty is acknowledged in the related technical literature to be the major reason for not meeting production expectations \cite{baker1998resource,asad2012optimal}. Given its substantial impact on the financial outcome of mining operations, this paper focuses on dealing with the uncertainty in metal content within a mineral deposit being mined. For the stochastic variables of the stockpile blending problem, we introduce chance-constrained programming here to tackle the uncertainty of material grades. Chance-constrained optimization problems~\cite{Charnes,Miller} whose resulting decision ensures the probability of complying with the constraints and the confidence level of being feasible to have received significant attention in the literature. Chance-constraint programming has been widely applied in different disciplines for optimization under uncertainty~\cite{Uryasev}. For example, chance-constraint programming has been applied in analog integrated circuit design~\cite{McConaghy}, mechanical engineering \cite{Mercado}, and other disciplines \cite{liu,poojari}. However, so far, chance-constraint programming has received little attention in the evolutionary computation literature~\cite{Zhang}.

It is difficult for MIP to tackle such a continuous optimization problem containing the nonlinear constraints. To address this challenge, this paper proposes two repair operators to tackles the complex constraints. Follow the paper \cite{Yue19}, we present the surrogate functions of the chance constraints by using Chebyshev's inequality. Furthermore, a well-known evolutionary algorithm, the Differential Evolution (DE) algorithm is introduced to solve the stockpile blending problem. Recently, evolutionary algorithms have received much attention in solving large-scale optimization problems and multi-dimensions problem. The DE algorithm is a simple and effective evolutionary algorithm used to solve global optimization problems in a continuous domain \cite{neri2010recent,pham2011comparative}. The DE and its variants have been successfully applied to solve numerous real-world problems from diverse domains of science and engineering \cite{das2010differential,neri2010recent}. This paper investigates the use of the DE algorithm combining the two repair operators for solving the problem. Then we compare the impactas of different chance constraints on the objective value.

The rest of the paper is organized as follows. In the next section, we present the model of the stockpile blending problem and a decision variable normalized operator for the continuous decision variables as well as a duration repair operator. After that, the chance constraints model and the surrogate functions of the chance constraints are presented in Section \ref{sec:chanceModel}. Following, we describe the approach we used to solve the problem and the fitness function of the algorithm. We set up experiments and investigate the performance of the different fitness functions in Section \ref{sec:experiment}. We conclude with Section \ref{sec:conclusion}.

\section{Deterministic model}

In this section, we present nonlinear formulations of the stockpiles blending problem with a deterministic setting. In reality, some processes such as the chemical process in the concentrate production progress are highly complex to model because it is influenced by many factors, some of which include the mineralogy of the ore, particle size of milled material, temperature, and chemical reactants available in the process. The information of these variables was not available to us, therefore within this study to recovery factors of all materials from the chemical processing stage and the copper percentage within the produced copper concentrate is assumed to be constant throughout the stockpiles blending and production schedule.  

We first introduces notation as follow, and then provide the math. We use the term "material" to include ore, i.e., rock that contains sufficient minerals including metals that can be economically extracted and to include waste, and we use chemical symbol represent the corresponding material, i.e., Cu denotes Copper, Fl denotes Flerovium.

\subsection{Notation}
\label{sec:notation}

\begin{table}[th]
    \centering
  \scriptsize
  \scalebox{1.0}{
 \makebox[\linewidth][c]{
\tabcolsep=0.5cm
    \begin{tabular}{ll}
   \multicolumn{2}{l}{\textbf{Indices and sets:} } \\
     $s\in \mathcal{S}$  & stockpiles; $1,\ldots,S$\\
       $p\in \mathcal{P}$ & parcels; $1,\ldots,P$\\
      $o$ & material; $\{Cu,Ag,Fe, Au,U, Fl, S\}$\\
      $m \in \mathcal{M} $ & month; $1,\ldots,m$ \\
           \\
      \multicolumn{2}{l}{\textbf{Decision variables:} }\\
       $x_{ps}$ & fraction of parcel $p$ claimed from stockpile $s$ \\
       $t_p$ & produce time (duration) for parcel $p$ \\
        $w_p$: & tonnage of parcel $p$\\
         $\theta_{ps}$: & tonnage stores in stockpile $s$ after providing material to parcel $p$\\
    $c_p$: & Cu tonne in parcel $p$\\
    $g^o_p$: & grade of material $o$ in parcel $p$\\
    $ \tilde{g}^o_{ps}$: & grade of material $o$ in stockpile $s$ when proving parcel $p$\\
    $k_p$: & tonne concentrate of parcel $p$\\
    $r^{Cu}_p$: &  Cu recovery of parcel $p$\\
    $r^{Fl}_p$: &  Fl recovery of parcel $p$\\
      \\
    \multicolumn{2}{l}{ \textbf{Parameters: }} \\
    $T_p^m$:  & binary parameter, if $T^m_p=1$, parcel  $p$\\ & is the first parcel need to prepare in month $m$, if $T^m_p=0$ otherwise\\
     $\delta$: & discount factor for time period\\ 
  $\tilde{\phi}$: & factor in chemical processing stage \\
 $\phi^{Au}$: & factor of Au in chemical processing stage \\
  $\phi^{U}$: & factor of U in chemical processing stage \\
  $\phi^{Fe}$: & factor of Fe in chemical processing stage \\
  $\phi^{Cu}$: & factor of Cu in chemical processing stage \\
  $(\gamma_1,\gamma_2)$ : & factor of Cu percentage within the produced Cu concentrate \\
  $\mu^{Fl}$: & factor of Fl recovery \\
  $\mu^{U}$ : & factor of U recovery \\
  $(\mu^{Cu}_1, \mu^{Cu}_2)$ : & factor of Cu recovery \\
  $D^m$:   & duration of month $m$ \\
  $H^m_s$:   & tonnage of material hauled to stockpile $s$ in month $m$\\
  $G^{om}_s$: & grade of material $o$ that shipping to the stockpile $s$ in month $m$\\
  $K_p$: & expected tonne concentrate of parcel $p$\\
  $R^{Fl}_p$ : & upper threshold of Fl recovery of parcel $p$\\
  ${Cu}_p$ : & lower threshold of Cu grade of parcel $p$\\
  $N_m$: & number of planning parcels in month $m$\\
\end{tabular}}}
  \label{tab:notation}%
\end{table}%

\subsection{Model with deterministic setting}

\begin{align}
  Obj:   \max \sum_{p\in \mathcal{P}} c_p = max \sum_{p\in \mathcal{P}} \left(w_p g^{Cu}_p r^{Cu}_p \right) 
    \label{obj:function}
\end{align}

\begin{align}
s.t. \sum_{\sum_{m=1}^{m-1}N_m +1 \leq p \leq \sum_{m=1}^m N_m} t_p \leq D^m 
\label{con:duration}
\end{align}

\begin{align}
 \sum_{s\in\mathcal{S}} x_{ps}=1 \qquad \forall p \in \mathcal{P} 
 \label{con:variables}
\end{align}

\begin{align}
    r^{Cu}_p = \mu^{Cu}_1 \frac{g^{Cu}_p}{g^S_p} +\mu^{Cu}_2 \qquad \forall p \in \mathcal{P}
    \label{con:curecovery}
\end{align}

\begin{align}
     g^o_p = \sum_{s\in \mathcal{S}} x_{ps} \tilde{g}^o_{ps} \qquad \forall p \in \mathcal{P} 
     \label{con:parcelgrade}
\end{align} 

\begin{align}
    w_p = & \delta t_p[\tilde{\phi} +(\phi^{Au}\log g^{Au}_p) + (\phi^{U}\log g^{U}_p)  \nonumber\\
    & -  (\phi^{Fe}\log g^{Fe}_p) + (\phi^{Cu}\log g^{Cu}_p)] 
    \label{con:parceltonne}
\end{align}

\begin{align}
    k_p= \frac{c_p}{\gamma_1 \frac{g^{Cu}_p}{g^{S}_p}+\gamma_2}
    \label{con:concentrate}
\end{align}

\begin{eqnarray}
  \tilde{g}^o_{ps} =
\begin{cases}
\frac{\tilde{g}^o_{(p-1)s}\cdot \theta{(p-1)s}+ G^{om}_s \cdot H^m_s}{\theta{(p-1)s}+H^m_s}  &   \textit{if $T_p^m=1$}
  \\
 \tilde{g}^o_{(p-1)s} &  otherwise
\end{cases} 
\label{con:stockpilegrade}
\end{eqnarray}

\begin{eqnarray}
\theta_{ps} =
\begin{cases}
\theta_{(p-1)s}+H^m_s - x_{ps} \cdot w_p  &   \textit{if $T_p^m=1$ } \\
\theta_{(p-1)s}- x_{ps} \cdot w_p  &  otherwise
\label{con:stockpilestorage}
\end{cases} 
\end{eqnarray}

\begin{align}
    g^{Cu}_p \geq {Cu}_p \qquad \forall p\in \mathcal{P}
    \label{con:cugradebound}
\end{align}

\begin{align}
  ( K_p-1)  \leq k_p \leq (K_p+1) \qquad \forall p\in \mathcal{P}
  \label{con:concentratebound}
\end{align}

\begin{align}
  \mu^{Fl} g^{Fl}_p   \leq R^{Fl}_p \qquad \forall p \in \mathcal{P}
  \label{con:Flrecoverybound}
\end{align}

The objective function (\ref{obj:function}) aims to maximize the sum of Cu volume of all parcels, which is obtained by the tonnage of parcels multiply the Cu grade, and multiple the Cu recovery. Constraint (\ref{con:duration}) forces the sum of duration of the parcels that planned into the same month less than the available duration of this month. Constraint (\ref{con:variables}) ensures that the sum of the decision variables for the same parcel is equal to 1. Function (\ref{con:curecovery}) denotes the simplified calculation of Cu recovery of parcels, and function (\ref{con:parcelgrade}) calculates the material grades of parcels. Function (\ref{con:parceltonne}) express the simplified calculation of parcel tonne which is a component in objection function. Function (\ref{con:concentrate}) shows the simplify version of how to calculate the tonne concentrate of parcels.

Constraint (\ref{con:stockpilegrade}) enforces material grade balance for stockpiles when providing material to parcels. Constraint (\ref{con:stockpilestorage}) enforces inventory balance when providing material to parcels.  Constraint (\ref{con:cugradebound}) forces the Cu grade of parcels to less than or equal to the given lower bound of the Cu grade. Constraint (\ref{con:concentratebound}) forces the value of tonne concentrate of each parcel is no more or less than the expected tonne concentrate by one. Constraint (\ref{con:Flrecoverybound}) ensures the Fl recovery of each parcel less than the bound given in advance.

\begin{algorithm}[t]
\caption{Decision variables normalized approach}
\label{alg:variablesapproach}
\KwIn{Decision vector $X_j=\{x_{1j},x_{2j}..,x_{Ij}\}$}
$a = \sum_{i \in I }x_{ij}$\;
\For {$i=1$ to $I$}{
 $x_{ij} = \frac{x_{ij}}{a}$\;
}
\Return the normalized decision variables.
\end{algorithm}

The stockpile blending problem is a non-linear optimization problem in the continuous search space. To tackle the constraint (\ref{con:variables}), we introduce a decision variables normalized approach (cf. Algorithm \ref{alg:variablesapproach}) to force  solutions match the constraint. This approach first calculates the sum of the variables of each parcel separately,  then each variable of the same parcel is divided by the corresponding sum. It shows the significant importance of applying this approach to a generated decision vector to meet the constraint.

Due to the complex constraint (\ref{con:concentratebound}) is too tight to construct feasible solutions, we develop a repair operator to address this problem. As shown in function (\ref{con:concentrate}), the value of tonne concentrate of parcels is related to the duration of this parcel and the material grades of the parcel. Meanwhile, referring to equation (\ref{con:parcelgrade}), material grades of the parcel are directly calculated by decision variables. Therefore, with the fixed decision variables of a parcel, the real tonne concentrate of this parcel is affected by the duration of this parcel. We present a duration repair operator (cf. Algorithm \ref{alg:durationfix}) which uses a binary search process to convert an infeasible solution into a solution without violating constraint (\ref{con:concentratebound}). 

\begin{algorithm}[t]
\caption{Duration repair operator}
\label{alg:durationfix}
\KwIn{$X\in (0,1)^{I\cdot J}$, $i\in \{1,..,I\}$, $j\in\{1,..,J\}$; parameter $\zeta$; available duration $\mathit{D}$} \
\KwOut{ parcel duration: $\mathit{d}\in \{0,\mathit{D}\}$}\
initialization: $\underline{d}=0$, $\overline{d}=\mathit{D}$, $d\in\{0,\mathit{D}\}$ , $k=\zeta \cdot d$ \
    \While{$d \in \{0, \mathit{D}\}$ and $k \notin \{K-1,K+1\}$ }{
    \If{$k>K+1$}{
    $d := (d+\underline{d})/2$\;
    
    $k := \zeta \cdot d$\;
    
    \If{$k>K+1$}{
    $\overline{d}=d$\;
    }\Else{
    $\underline{d}=d$\;
    }
    }
    \ElseIf{$k < K-1$}{
    $d := (d+\overline{d})/2$}\;
    $k := \zeta \cdot d$\;
    \If{$k>K+1$}{
    $\overline{d} :=d$\;
    }\Else{$\underline{d} :=d$}
    }
\Return \textit{the duration corresponding to solution $X$}
\end{algorithm}

Since the time complexity of the binary search is $\log n$ where $n$ denotes the length of the search space in the beginning. In our problem, the duration of each parcel can not exceed the total available duration of the month. The run-time of the duration repair operator for one parcel is $\log d$ in the worst case where $d$ denotes the total available duration of the current month.

\section{Model with chance constraints}
\label{sec:chanceModel}

In real-world mining engineering problems, the material grades are estimated by some tools, and in the research of mining scheduling problems, researchers treat the stochastic material grades as constant by using the expected values. In this paper, we are the first to discuss the influences of stochastic material grades on the objective value of the stockpile blending problem with chance constraints. Due to the complexity of the problem, we reformulated the constraints (\ref{con:cugradebound}) and (\ref{con:Flrecoverybound}) to chance-constrained. Chance-constrained programming is a competitive tool for solving optimization problems under uncertainty. The main feature is that the resulting decision ensures the probability of complying with constraints, i.e. the confidence of being feasible. Thus, using chance-constrained programming the relationship between profitability and reliability can be quantified.

\subsection{The formulation of the chance constraints}

First, we define additional notation as follow.
\begin{table}[h]
 \centering
  \scalebox{1.0}{
  \makebox[\linewidth][c]{
  \tabcolsep=0.5cm
\begin{tabular}{ll}
  $\alpha_{Cu}$:   &  confidence of Cu grade chance constraint \\
   $\alpha_{Fl}$:   &  confidence of Fl recovery chance constraint \\
 \end{tabular}}}
\end{table}

Then, the new chance constraints are
\begin{align}
    Pr\{g^{Cu}_p \geq Cu_P\} \geq \alpha_{Cu},
    \label{chance:coppergrade}
\end{align}
and
\begin{align}
    Pr\{\mu^{Fl}g^{Fl}_p \leq R_{p}^{Fl}\} \geq \alpha_{Fl}.
    \label{chance:Flrecovery}
\end{align}
Constraints (\ref{chance:coppergrade}), (\ref{chance:Flrecovery}) force the confidence of ensuring the constraint are greater than or equal to the corresponding given bound. 

We use Chebyshev's inequality to construct the available surrogate that translates to a guarantee on the feasibility of the chance constraint imposed by the inequalities. Firstly, we use Chebyshev's inequality to reformulate the chance constraints. The inequality has utility for being applied to any probability distribution with known expectation and variance. Therefore, we assume the stochastic material grades discussed in this paper are all estimated with given expected values and corresponding variances. Note that Chebyshev's inequality automatically yields a two-sided tail bound, there is a one-sided version of Chebyshev's inequality named Cantelli's inequality.

\begin{theorem}[Cantelli's inequality]
\label{thm:cheb}
  Let $X$ be a random variable with $Var[X]>0$. Then for all $\lambda>0$, 
  \begin{equation}
    P_r\{X\geq E[X] +\lambda\sqrt{Var[X]}\}\leq \frac{1}{1+\lambda^2}.
    \label{the:Canplus}
  \end{equation}
  
   \begin{equation}
    P_r\{X\leq E[X] -\lambda\sqrt{Var[X]}\}\leq \frac{1}{1+\lambda^2}.
    \label{the:Canminus}
  \end{equation}
\end{theorem}

We assume the material grades in the stockpiles are independent of each other, each grade corresponding expectation $a^{om}_{s}$ and variance $\sigma_s^{2om}$. Therefore, the expected material grades of stockpiles can be denoted as
\begin{eqnarray}
  E(\tilde{g}^o_{ps}) =
\begin{cases}
\frac{\tilde{g}^o_{(p-1)s}\cdot \theta_{(p-1)s}+ a^{om}_s \cdot H^m_s}{\theta{(p-1)s}+H^m_s}  &   \textit{if $T_p^m=1$}
  \\
 E(\tilde{g}^o_{(p-1)s}) &  otherwise.
\end{cases} 
\nonumber
\end{eqnarray}
Furthermore, the variance of the material grades are
\begin{eqnarray}
  Var(\tilde{g}^o_{ps})=
\begin{cases}
\left(\frac{\theta_{(p-1)s}}{\theta{(p-1)s}+H^m_s}\right)^2 Var(\tilde{g}^o_{(p-1)s})  +\left(\frac{H^m_s}{\theta{(p-1)s}+H^m_s}\right)^2 \sigma_s^{2om} &   \textit{if $T_p^m=1$}
  \\
Var(\tilde{g}^o_{(p-1)s}) &  otherwise.
\end{cases} 
\nonumber
\end{eqnarray}

Let $g^{Cu}_p = \sum_{s\in \mathcal{S}} x_{ps}\tilde{g}^{Cu}_{ps}$ be the Cu grade of parcel $p$ of a given solution $X=\{x_{p1},..,x_{ps},..,x_{p\mathbf{S}}\}$, and $$E[g^{Cu}_p ]=\sum_{s\in \mathcal{S}}x_{ps}E(\tilde{g}^{Cu}_{ps})$$ denotes the expected Cu grade of parcel $p$ of the solution derived by linearity of expectation, $$Var[g^{Cu}_p]=\sum_{s\in \mathcal{S}} (x_{ps})^2 Var(\tilde{g}^{Cu}_{ps})$$ denotes the variance of Cu grade of parcel $p$. To match the expression of the Cantelli's inequality (\ref{the:Canminus}), we set $$Cu_p=E[g^{Cu}_p]-\lambda\sqrt{Var[g^{Cu}_p]}$$ and have $$\lambda =\frac{E[g^{Cu}_p]-Cu_p}{\sqrt{Var[g^{Cu}_p]}}$$ for each parcel, then we have a formulation to calculate the upper bound of the chance constraint (\ref{chance:coppergrade}) as follows.

\begin{align}
    Pr\{g^{Cu}_p \leq Cu_p\} \leq \frac{Var[g^{Cu}_p]}{Var[g^{Cu}_p]+(E[g^{Cu}_p]-Cu_p)^2} \leq (1-\alpha_{Cu})
    \label{chance:cu}
\end{align}

Furthermore, let $ r^{Fl}_p=\mu^{Fl}\sum_{s\in \mathcal{S}}x_{ps}\tilde{g}^{Fl}_{ps} $ 
be the FL recovery of parcel $p$. Let $$E[r^{Fl}_p ]=\mu^{Fl}\sum_{s\in \mathcal{S}}x_{ps}E(\tilde{g}^{Fl}_{ps})$$
denotes the expectation of Fl recovery, and $$Var[r^{Fl}_p ]=\sum_{s\in \mathcal{S}} (\mu^{Fl} x_{ps})^2 Var(\tilde{g}^{Fl}_{ps}),$$ 
is the variance of FL recovery of parcel $p$ with solution $X=\{x_{p1},..,x_{ps},..,x_{p\mathcal{S}}\}.$

To match the expression of the Cantelli's inequality (\ref{the:Canplus}), we set $$R_{p}^{Fl}=\mu^{Fl}E[r_p^{Fl}]+\lambda\sqrt{Var[r^{Fl}_p]}$$ and have
$$\lambda = \frac{R_p^{Fl}-\mu^{Fl}E[g^{Fl}_p]}{\sqrt{Var[g^{Fl}_p]}}$$ for each parcel, then we have a formulation to calculate the upper bound of the chance constraint (\ref{chance:Flrecovery}) as follows.
\begin{align}
    \resizebox{0.9\hsize}{!}{$Pr\{\mu^{Fl}g^{Fl}_p\geq R^{Fl}_p\} \leq \frac{Var[g^{Fl}_p]}{Var[g^{Fl}_p]+(R_p^{Fl}-\mu^{Fl}E[g^{Fl}_p])^2} \leq (1-\alpha_{Fl})$}
    \label{chance:fl}
\end{align}

Now, we obtain the surrogate functions of the chance constraints. In the next section, we present the approach for solving the stockpile blending problem with chance constraints.

\section{Approaches for the stockpile blending problem with chance constraints}
\label{sec:approach}

In this section, we present the fitness functions for the differential evolution (DE) algorithm which has been proved successfully used in solving the optimization problem in continuous space.

\subsection{Fitness function for deterministic setting}

We start by designing a fitness function for the deterministic setting model that can be used in the DE algorithm. The fitness function $f$ for the approach needs to take all constraints into account. The fitness function of a solution $X$ is defined as follows.
\begin{align}
    f(X) = \left(u(X),v(X),w(X),q(X),g(X), O(X) \right)
    \label{fit:deter}
\end{align}

\begin{align}
    & u(X)=\sum_{p\in \mathcal{P}} \max \{\left|K_p-k_p \right|,1\} \nonumber \\
    & v(X)= \max\{\sum_{p\in \mathcal{P}}t_p - D^m , 0\} \nonumber \\
    & w(X)=\min \{\sum_{p\in \mathcal{P}}\sum_{s\in \mathcal{S}}\theta_{ps},0\} \nonumber \\
    & q(X)=\sum_{p \in \mathcal{P}} \max\{Cu_{p}- g^{Cu}_p,0\} \nonumber \\
    & g(X)=\sum_{p \in \mathcal{P}} \max\{r^{Fl}_p-R^{Fl}_p,0\} \nonumber \\
    & O(X)= \sum_{p\in \mathcal{P}} c_p. \nonumber
\end{align}

In this fitness function, the components $u,v,q,g$ need to be minimize while $w$ and $O$ maximized, and we optimize $f$ in lexicographic order. For the stockpile blending problem, any infeasible solution can at least violate one of the above constraints. Then, among solutions that meet all constraints, we aim to maximize the objective function. Formally, we have
\begin{align*}
&f(X) \succeq f(Y) \\
\textbf{iff} \  & u(X) <u(Y) \  \textit{or} \\
&{u(X)=u(Y)\wedge v(X)<v(Y)} \  \textit{or}\\
& \{u,v\} \text{are equal} \wedge w(X)>w(Y)  \  \textit{or}\\
& \{u,v,w\} \text{are equal} \wedge q (X)<q(Y)  \  \textit{or} \\
& \{u,v,w,q\} \text{are equal} \wedge g(X)<g(Y) \textit{or} \\
&\{u,v,w,q,g\} \text{ are equal}  \wedge O(X)>O(Y), 
\end{align*}
When comparing two solutions, the feasible solution is preferred in a comparison between an infeasible and a feasible solution. Between two infeasible solutions that violated the same constraint, the one with a lower degree of constraint violation is preferred.

\subsection{Fitness function of the problem with chance constraints}

Now, we design the fitness function for the stockpile blending problem with chance constraints. In this paper, we investigate the effectiveness of chance constraints on the objective value. We first reformulate the components $q$ and $g$ of the function (\ref{fit:deter}) with chance constraints (\ref{chance:cu} and \ref{chance:fl}) as follow,
\begin{align}
    q'(X) =\sum_{p\in \mathcal{P}} max\left\{P_r\{g^{Cu}_p \leq Cu_p\}-(1-\alpha_{Cu}),0\right\}\\
    g'(X) =\sum_{p\in \mathcal{P}} max\left\{P_r\{\mu^{Fl}g^{Fl}_p \geq R_p^{Fl}\}-(1-\alpha_{Fl}),0\right\}
\end{align}
where $q'$ and $g'$ need to be minimized. 

To distinguish the influence of each chance constraint, we design three fitness functions where the two functions consider the chance constraints separately, and the other one uses the combination of components.
\begin{align}
    f'(X) = \left(u(X),v(X),w(X),q'(X),g(X), O(X) \right)
    \label{fit:Cuonly}\\
    f''(X) = \left(u(X),v(X),w(X),q(X),g'(X),O(X) \right)
    \label{fit:flonly}\\
   f'''(X) = \left(u(X),v(X),w(X),q'(X),g'(X), O(X) \right)
   \label{fit:combina}
\end{align}

\subsection{Differential evolution algorithm}

\begin{algorithm}[t]
\caption{Differential evolution algorithm}
$t \leftarrow 1$, initialize $\mathbf{P}^t =\{X_1^t,..,X_{NP}^t\}$ randomly \;

\While{stopping criterion not met}{
    \For{$i\in\{1,..,NP\}$}{
     $R\leftarrow$ A set of randomly selected indices from $\{1,..,NP\} \setminus \{i\}$ \;
     $V^t_i \leftarrow$ mutation $(P^t,R,F)$ \;
     $j_{rand} \leftarrow$ A randomly selected number from $\{1,..,n\};$ \
     $U^t_i \leftarrow$ crossover$(X^t_i, V^t_i,C,j_{rand});$ \
     }
     \For{$i\in\{1,..,NP\}$}{
     \If{$f(U^t_i) \succeq f(X^t_i)$}{
     $X^{t+1}_i \leftarrow U^t_i$\;
     }
     \Else {
     $X^{t+1}_i \leftarrow X^t_i$ \;}
     }
     $t\leftarrow t+1$ \;
     }
\label{alg:de}
\end{algorithm}

Algorithm (\ref{alg:de}) shows the overall procedure of the basic DE algorithm. DE is usually initialized by generating a population of $NP$ individuals. For each $i\in\{1,..,NP\}$, $X^t_i$ is the $i$-th individual in the population $\mathbf{P}^t$. Each individual represents a $d$-dimensional solution of a problem. For each $j\in\{1,..,d\}$, $X_{ij}^t$ is the $j$-th element of $X^t_i$.

After the initialization of $\mathbf{P}^t$, the following steps are repeatedly performed until a termination condition is satisfied. For each $X^t_i$, the scale factor $F>0$ which controls the magnitude of the mutation, and the crossover rate $Cr\in[0,1]$ which controls the number of elements inherited from $X^t_i$ to a trail vector $U^t_i$ are constants and given in advance. 

A set of parent indices $R=\{r_1,r_2\}$ are randomly selected from $\{1,..,n\}\setminus\{i\}$ such that they differ from each other. For each $X^t_i$, a mutant vector $V^t_i$ is generated by applying a mutation to $X^t_{r1},X^t_{r2}$. There are many mutation strategies that have been proposed in the literature \cite{das2010differential}. Here, we use the $DE/target-to-best/1$ strategy shown as follows, which is one of the most efficient strategies. 
\begin{align}
    V^t_i= X^t_i+F(X^t_{best}-X^t_i) +F(X^t_{r1}-X^t_{r2}),
\end{align}
where $X^t_{best}$ denotes the best individual in the current population. 

After the mutant vector $V^t_i$ has been generated for each $X^t_i$, a trail vector $U^t_i$ is generated by applying crossover to $X^t_i$ and $V^t_i$. The scheme of the crossover can be outlined as 
\begin{eqnarray}
U^t_{ij} =
\begin{cases}
V^t_{ij} &   \textit{if}\  (rand_{i,j}[0,1] \leq Cr \ \textit{or}\  j=j_{rand}) \\
X^t_{ij}  &  otherwise
\label{DE:crossover}
\end{cases} 
\end{eqnarray}
where $rand_{i,j}[0,1]$ is a uniformly distributed random number, which is called a new for each $j$-th element of the $i$-th parameter vector. $j_{rand}\in \{1,..,n\}$ is a randomly chosen index, which ensures that $U^t_i$ gets at least one element from $V^t_i$. It is instantiated once for each vector per generation.

After the trial vector, $U^t_i$ has been generated for each parent individual, the next step called selection which determines whether the target or the trailing vector survives to the next generation. The selection operation is described as 
\begin{eqnarray}
X^{t+1}_{i} =
\begin{cases}
U^t_{i} &   \textit{if}\  f(U^t_i)\succeq f(X^t_i)  \\
X^t_{ij}  &  otherwise
\label{DE:selection}
\end{cases} 
\end{eqnarray}
according to the fitness function.

\section{Experimental investigation}
\label{sec:experiment}
In this section, we examine the solution quality associated with different fitness functions. Due to business security, we are not able to investigate the proposed approach in real-data instances. Therefore, we first design the benchmark of the stockpile blending problem. Afterward, we compare the results obtained by using different fitness functions of the instances. Furthermore, considering the complexity of the problem with chance constraints, the instances we discussed in this section only contain one month schedule.

\subsection{Experimental Setup}

Table \ref{tab:paramRange} lists the intervals of input parameters mentioned in Section \ref{sec:notation}. The three instances we evaluated in this paper are created by randomly generated value of parameters from their intervals (see Table \ref{tab:paramRange}), we attach the parameters of these instances in the appendix. The randomly generated numbers are the expected values of material grades, and the deviation of material grades are set equal to $0.01$ multiply the expectation. Let $\alpha_{Cu}=\{0.999,0.99,0.9\}$ and $\alpha_{Fl}=\{0.999,0.99,0.9\}$. Base on this arrangement, we compare the performance of the DE algorithm with fitness functions (Eq. \ref{fit:deter}, \ref{fit:Cuonly}, \ref{fit:flonly}, \ref{fit:combina}) on the stockpile blending problem.

\begin{table}[t]
\centering
\caption{General information about the ore and processing parameters}
  \scriptsize
  \scalebox{1.0}{
 \makebox[\linewidth][c]{
    \begin{tabular}{ll}
     \toprule
   \text{Description} & \text{Values or Value range}\\
     \midrule
     Number  of parcel & stockpiles; $\{3,4,5\}$\\
     Number of Stockpile & $7$\\
     Duration of month & ${29,30,31}$ \\
     Discount factor for time period ($\delta$) & $0.98$\\
     Factor in chemical processing stage ($\tilde{\phi}$) & $[1000,2000]$ \\
     Factor of Au in chemical processing stage ($\phi^{Au}$) & $[200,300]$ \\
     Factor of U in chemical processing stage ($\phi^{U}$) & $[300,400]$ \\
     Factor of Fe in chemical processing stage ($\phi^{Fe}$) & $[560000,570000]$ \\
     Factor of Cu in chemical processing stage ($\phi^{Cu}$) & $[6000000,7000000]$ \\
     Factor of Cu percentage within the produced Cu concentrate $(\gamma_1,\gamma_2)$ & $([5,10],[30,40])$ \\
     Factor of Fl recovery ($\mu^{Fl}$) & $[0.05,0.15]$\\
     Factor of U recovery  ($\mu^{U}$) & $[0.5,0.9]$\\
     Factor of Cu recovery $(\mu^{Cu}_1, \mu^{Cu}_2)$ & $([1.5,3.5],[0,10])$\\
     Tonnage of material hauled to stockpile & $[5000,1000000]$\\
     Cu grade & $[0.05,2.5]$\\
     Ag grade & $[1.0,4.0]$\\
     Fe grade &  $ [10.0,30.0]$\\
     Au grade & $[0.3,2.0]$\\
     U grade & $[30.0,400.0]$\\
     Fl grade & $[1200,4500]$\\
     S grade & $[0.15,1.0]$ \\
     Expected tonne concentrate of parcel ($K_p$) & $[10000,\infty]$\\
     Threshold of Fl recovery of parcel ( $R^{Fl}_p$) & $[1300,1500]$\\
     Threshold of Cu grade of parcel (${Cu}_p$) & $[0.5,1.5]$\\
   \bottomrule
   \end{tabular}}}%
\label{tab:paramRange}
\end{table}%

We then investigate the performance of the DE algorithms with different fitness functions described in Section \ref{sec:approach} and provide the results from $30$ independent runs with $10000$ generation and $10$ population for all instances. For a closer look, we report the average, best and worst solutions obtained by the algorithm in corresponding columns. We also evaluate the algorithm by success rate which is the percentage of success for the algorithm in obtaining valid solutions out of $30$ runs. 

\begin{table*}[t]
  \centering
  \caption{Fitness values obtained with single chance constraint}
  \scalebox{0.6}{
  \makebox[\linewidth][c]{
    \begin{tabular}{llrrrrrrrrr}
    \toprule
          &  &   Deterministic &    & \multicolumn{3}{l}{Cu Chance constraint $(\alpha_{Cu})$} &       & \multicolumn{3}{l}{Fl Chance constraint $(\alpha_{Fl})$ } \\
    Instance &       &     &  & 0.999 & 0.99  & 0.9   &       & 0.999 & 0.99  & 0.9 \\
    \midrule
    \multirow{4}{*}{1} & Mean  & 103603035.94 &  & 99319128.52 & 102724900.09 & 103206748.35 &       & 103117715.98 & 103340755.20 & 102753876.59 \\
          & Best &   110830487.20 & & 100434268.90 & 110489368.20 & 111221777.51 &       & 108460913.10 & 110593860.12 & 106976825.01 \\
          & Worst & 100025173.10 & & 98404158.16 & 99426947.62 & 99935646.96 &       & 99979453.63 & 98951340.30 & 99444063.80 \\
          &      Success rate & & & 0.166666667 & 1     & 1     &       & 1     & 1     & 1 \\
          \midrule
    \multirow{4}[0]{*}{2} &Mean  & 66339866.34 & & 64280794.53 & 65346088.50 & 65440690.91 &       & 65741114.11 & 66128957.43 & 64999603.10 \\
          &Best  & 69691652.87 &  & 66062865.43 & 70504938.14 & 69307609.02 &       & 70846603.30 & 69265095.80 & 67416065.20 \\
          & Worst  & 63401302.85 & & 62822049.78 & 61409848.38 & 62043167.43 &       & 62885220.60 & 63139531.31 & 62369471.32 \\
          &   Success rate    &  &  &0.3   & 1     & 1     &       & 1     & 1     & 1 \\
          \midrule
    \multirow{4}[0]{*}{3} &Mean  & 25706739.82  & & 25172345.85 & 25484058.55 & 25667396.95 &       & 25414602.50 & 25487090.60 & 25737554.80 \\
          &Best  & 26714591.61 &  & 25780112.93 & 27501228.19 & 26652385.99 &       & 26542657.00 & 26440088.30 & 27420748.21 \\
          &Worst &  25042412.92 & & 24939884.30 & 24517912.31 & 24338065.15 &       & 24301347.20 & 24502516.90 & 24675079.20 \\
          &   Success rate    & & & 0.166666667 & 0.733333333 & 0.8   &       & 0.83333333 & 0.7   & 1 \\
          \bottomrule 
    \end{tabular}}}%
  \label{tab:singleResults}%
\end{table*}%

\begin{table*}[t]
  \centering
  \caption{Fitness values obtained with two chance constraints}
   \scalebox{0.6}{
  \makebox[\linewidth][c]{
    \begin{tabular}{llrrrrrrrrrrrr}
    \toprule
    Instance &  & &\multicolumn{11}{c}{Combine Chance constraints} \\
     &  &   & \multicolumn{3}{c}{$\alpha_{Cu}=0.999$} &       & \multicolumn{3}{c}{$\alpha_{Cu}=0.99$} &       & \multicolumn{3}{c}{$\alpha_{Cu}=0.9$} \\
          \midrule
    $\alpha_{Fl}$ & & & 0.999 & 0.99  & 0.9   &       & 0.999 & 0.99  & 0.9   &       & 0.999 & 0.99  & 0.9 \\
    \midrule
    \multirow{4}[0]{*}{1} & Mean &  & 98787701.39 & 99241024.07 & 99492631.19 &       & 102600300.50 & 102276579.70 & 102682519.61 &       & 103388918.00 & 102934747.10 & 102493177.23 \\
       &Best  &    & 102031085.80 & 103755489.91 & 102599092.62 &       & 106167319.20 & 107035409.13 & 109936137.61 &       & 106369047.00 & 108285976.13 & 108115669.08 \\
       &Worst  &    & 96585053.09 & 97115231.16 & 97595258.99 &       & 99322445.27 & 98370524.81 & 100035474.00 &       & 100191353.05 & 99131474.70 & 99124405.82 \\
    &Success rate &        & 0.366666667 & 0.366666667 & 0.4   &       & 1     & 1     & 1     &       & 1     & 1     & 1 \\
          \midrule
    \multirow{4}[0]{*}{2} & Mean & & 65983255.56 & 63563990.36 & 64902632.44 &       & 65902068.31 & 65729892.86 & 65506695.73 &       & 65678975.14 & 65559860.50 & 65651773.33 \\
         &Best  &     & 69079010.41 & 65782360.39 & 69006540.86 &       & 69991791.36 & 69754992.24 & 70596782.19 &       & 69703718.21 & 68258615.00 & 70556638.81 \\
        &Worst  &      & 62870362.84 & 61877215.08 & 61963922.28 &       & 59672821.89 & 63038071.83 & 61823714.56 &       & 61900565.70 & 61618562.90 & 62959270.15 \\
         &Success rate   &     & 0.166666667 & 0.133333333 & 0.3   &       & 1     & 1     & 1     &       & 1     & 1     & 1 \\
          \midrule
    \multirow{4}[0]{*}{3}& Mean & & 25960917.39 & 25871636.05 & 25686417.52 &       & 25459522.87 & 25682180.87 & 25687541.01 &       & 25569238.21 & 25498602.00 & 25617280.63 \\
   & Best  &         & 25960917.39 & 26273134.25 & 25686417.52 &       & 26575050.89 & 26643851.74 & 26329733.68 &       & 26457335.14 & 26589310.80 & 26326709.05 \\
       & Worst  &     & 25960917.39 & 25473933.01 & 25686417.52 &       & 24495873.72 & 24826217.28 & 24668501.30 &       & 24693215.90 & 24721583.71 & 24549825.10 \\
      & Success rate  &      & 0.033333333 & 0.133333333 & 0.033333333 &       & 0.8   & 0.833333333 & 0.866666667 &       & 0.8   & 0.86666667 & 0.76666667 \\
          \bottomrule
    \end{tabular}}}%
  \label{tab:MultiResults}%
\end{table*}%

\subsection{Experimental Results}

We benchmark our approach with the combinations from the experimental setting described above. All experiments were performed using Java of version 11.0.1 and carried out on a MacBook with a 2.3GHz Intel Core i5 CPU. 

\begin{figure*}[h]
\centering
\subfigure[Instance 1]{
\includegraphics[width = 0.3\textwidth]{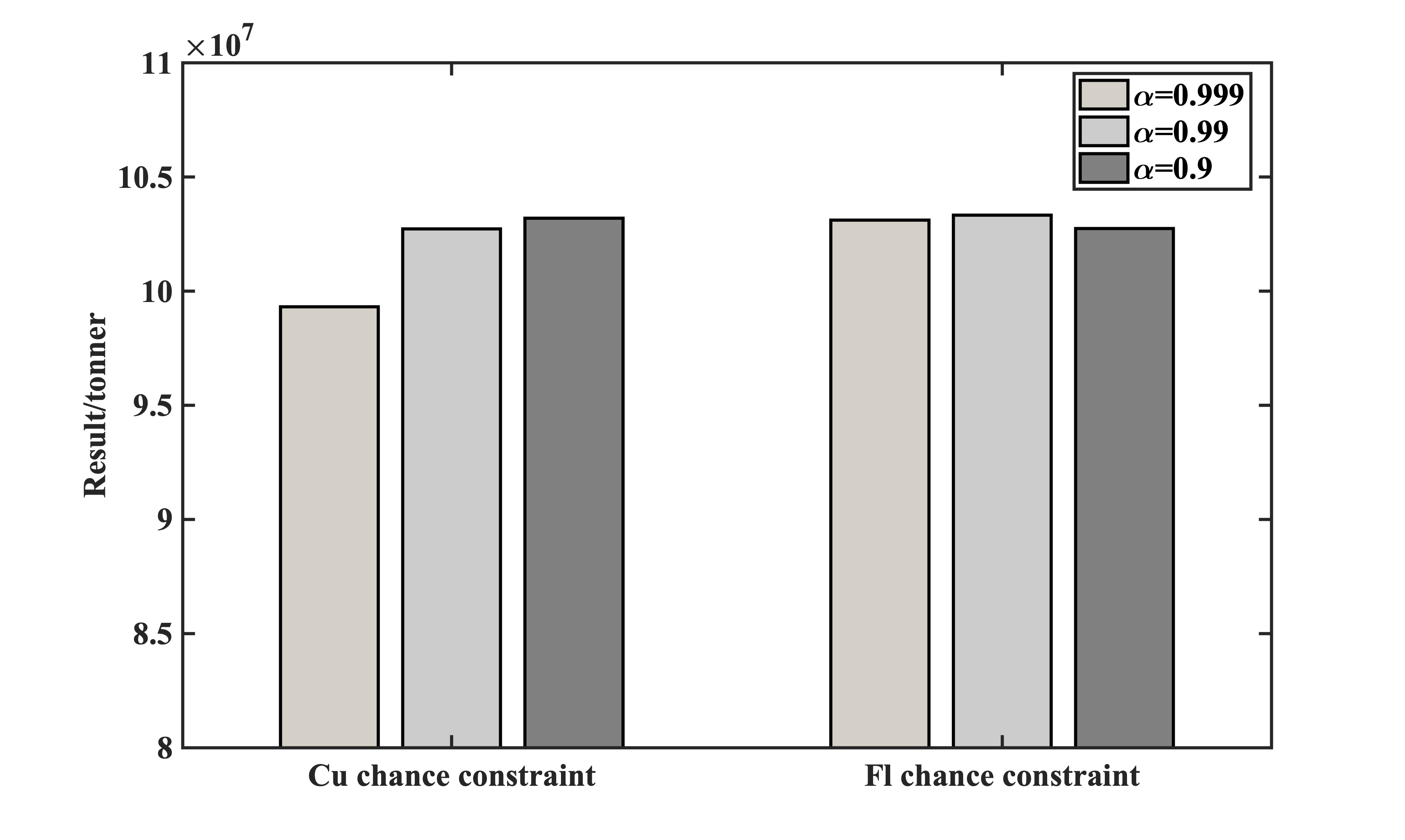}
}
\subfigure[Instance 2 ]{
\includegraphics[width = 0.3\textwidth]{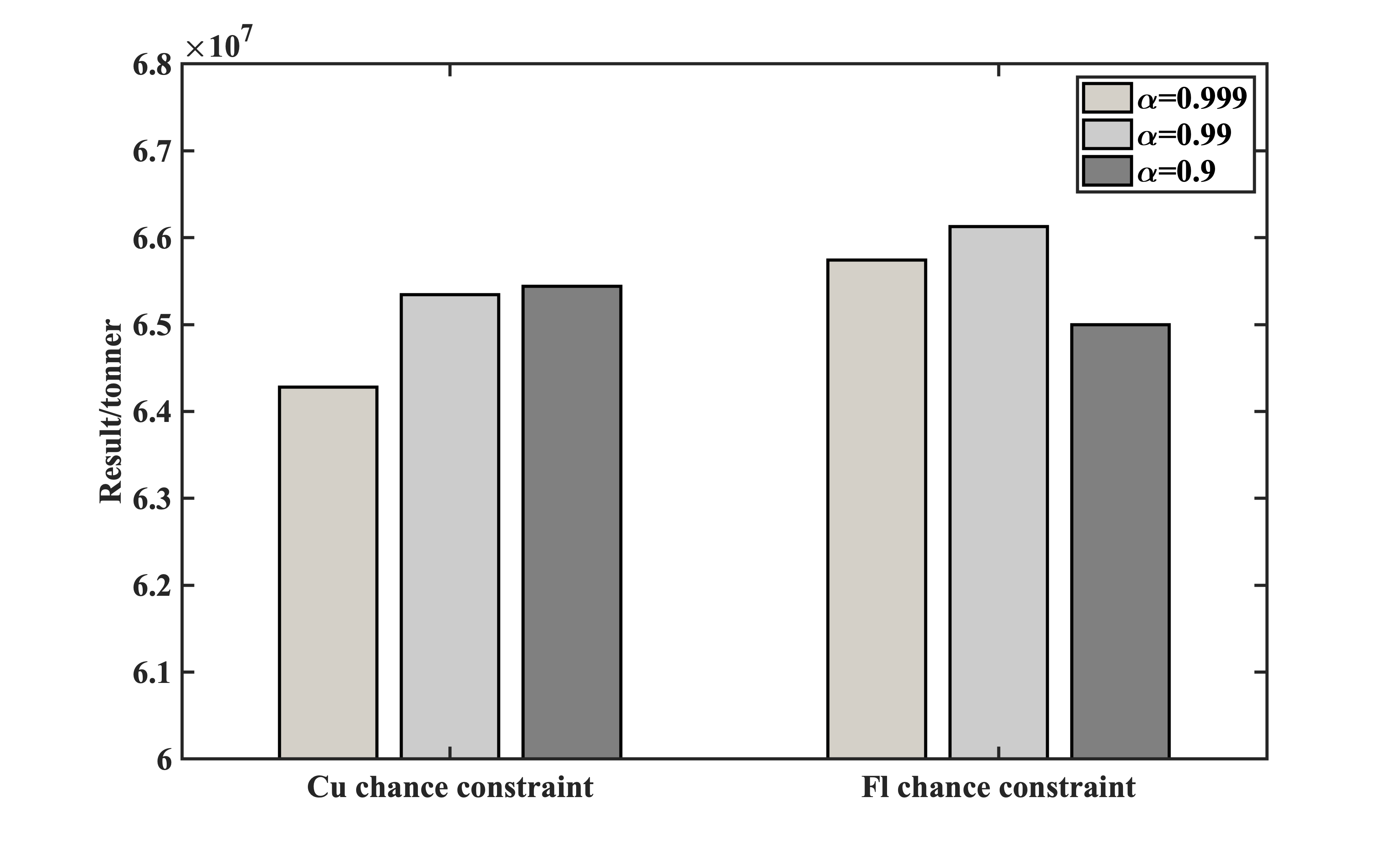}
}
\subfigure[Instance 3]{
\includegraphics[width = 0.3\textwidth]{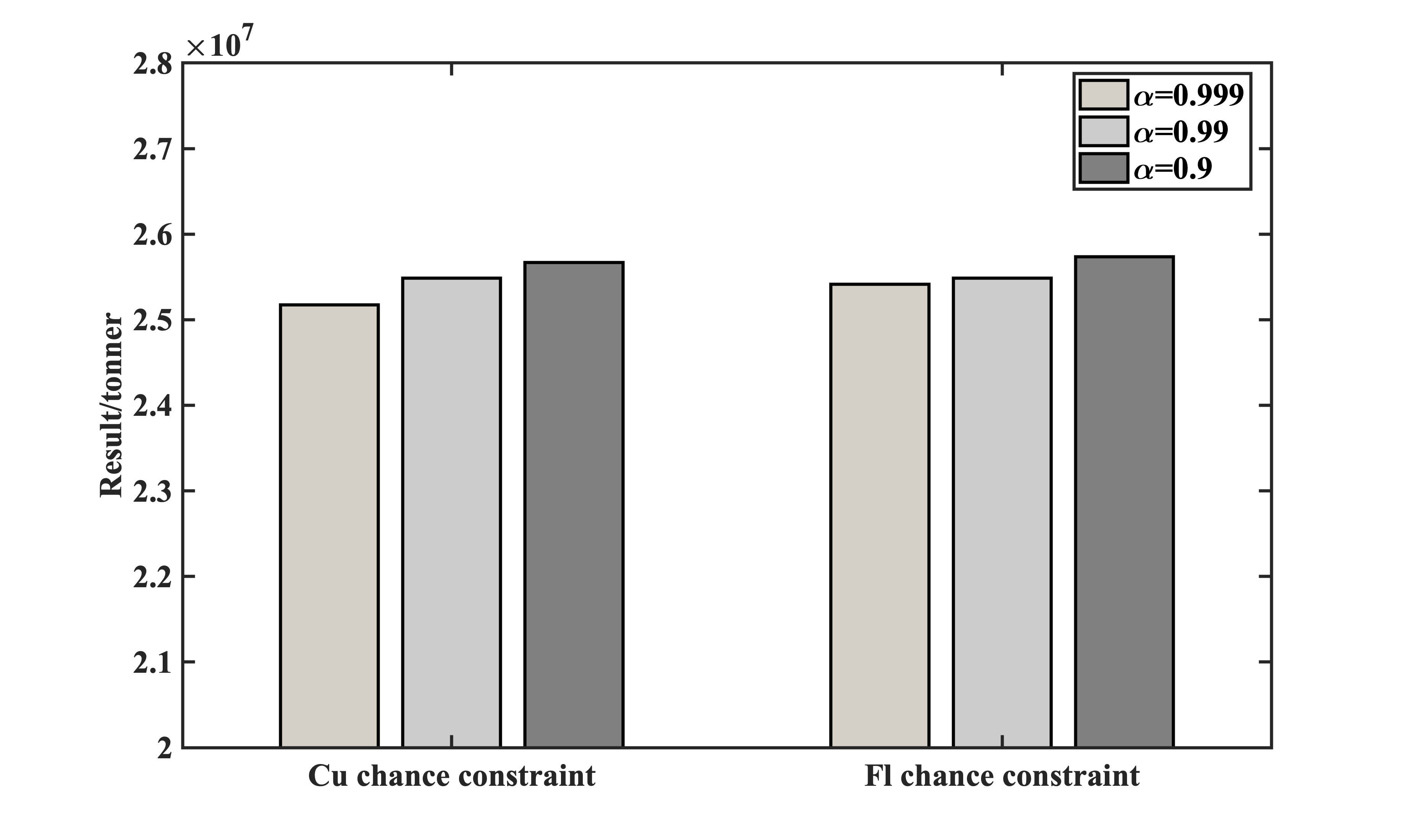}
}
\caption{Bar graph for DE algorithm with single chance constraint }

\label{fig:bargraph}
\end{figure*}

Table \ref{tab:singleResults} lists the results for the three instances with using fitness function (\ref{fit:deter},\ref{fit:Cuonly}) and (\ref{fit:flonly}) separately.  Figure \ref{fig:bargraph} shows the how the chance-constrained bound $\alpha_{Cu}$ or $\alpha_{Fl}$ affects the quality of the solutions. The bars in the graphs are corresponding to the solutions of instances combining with the confidence of chance constraint respectively, and the three bars in each group corresponding to the threshold of confidence $\{0.999,0.99,0.9\}$. Among others, we observe that results obtained by applying the fitness function (\ref{fit:Cuonly}) are significantly affected by the value of $\alpha_{Cu}$. The results show an increasing trend as the value of $\alpha_{Cu}$ decrease. However, by observing the bars in \textit{Fl chance constraint} group, the value of $\alpha_{Fl}$ does not influence the result when using the Fl chance constraint.

As can be seen from Table \ref{tab:singleResults}, the success rate shows significantly difference between using the fitness functions (\ref{fit:Cuonly}) and (\ref{fit:flonly}) for instance $1$ and $2$. When the confidence of the chance constraint (\ref{chance:coppergrade}) is tight such as $0.999$, the DE algorithm can not generate a pure feasible population in the last generation. While the confidence of the chance constraint (\ref{chance:Flrecovery}) does not influence the success rate of the algorithm. However, for instance $3$, which has four parcels into consideration and is the most complex instance in our study, the DE algorithm fails to obtain a feasible population in the last generation when the value of $\alpha_{Fl}$ is $0.999$.

Table \ref{tab:MultiResults} lists the results obtained by considering two chance constraints together, the fitness function (\ref{fit:combina}). For each instance, we investigate different parameters setting together with the different requirement on the chance constraints determined by $\alpha_{Cu}$ and $\alpha_{Fl}$. The results list in the columns with the same $\alpha_{Cu}$ shows that there is no significant difference between the solutions obtained by applying difference $\alpha_{Fl}$. Moreover, with the same $\alpha_{Fl}$, the object value increase while the $\alpha_{Cu}$ decrease.

Now, we compare the results obtained by using single chance constraint and combined chance constraints. Comparing the solutions list in the column \textit{Cu Chance constraint} and \textit{Fl Chance constraint} in Table \ref{tab:singleResults} against that of the combined chance constraint in the same value of $\alpha_{Cu}$ and $\alpha_{Fl}$ respectively. We find that for the same instance, the results obtained by applying a single chance constraint are better than the combined chance constraints which happened in most cases. One interesting finding is that the value of $\alpha_{Fl}$ does not show significant effects on the results in the experiments for results in Table \ref{tab:singleResults} and \ref{tab:MultiResults}. A possible explanation for this might be that the parameters of the instances are not reliable or match the real-world situation, which can indicate the malfunction of the constraint. This is an important issue for feature research that develops approaches to create a benchmark that more reliable or more close to the real-world situation for the stockpile blending problem.

\section{Conclusion}
\label{sec:conclusion}

In this paper, we consider the stockpile blending problem which is an important component in mine scheduling with the uncertainty in the geologic input data. We modeled the stockpile blending problem as a nonlinear optimization problem and introduced the chance constraints to tackle the stochastic material grades. We show how to incorporate a well-known probability tail, Chebyshev's inequality, into presenting the surrogate functions of the chance constraints. Furthermore, we designed the four fitness functions with considering different chance constraints. In our experiments, which have covered a variety of instances according to the parameters, we have observed that the confidence of the Cu chance constraint affects the results obtained by using the fitness function considering the Cu chance constraint and the fitness function with combined chance constraints. Due to the ineffectiveness of the confidence of the Fl chance constraint, for further studies, it could be interesting to deeply investigate the relationship between chance constraints. It would be also interesting to develop benchmarks for the stockpile blending problem with chance constraints as there is no available open access data-set.

\section{Acknowledgements}
This research has been supported by the SA Government through the PRIF RCP Industry Consortium

\bibliographystyle{abbrv}
\bibliography{main}

\newpage

\section{Appendix}
Table \ref{tab:ins1},\ref{tab:ins2} and \ref{tab:ins3} list the value of input parameters of instance 1, 2 and 3 separately.

\begin{table*}[htbp]
  \centering
  \caption{Parameters of Instance 1}
  \scriptsize
  \scalebox{0.8}{
 \makebox[\linewidth][c]{
    \begin{tabular}{lrrrrrrrr}
    \toprule
    \multicolumn{1}{l}{Number of parcels:} & 3     &       &       &       &       &       &       &  \\
    \multicolumn{1}{l}{Number of stockpiles:} & 7     &       &       &       &       &       &       &  \\
    \multicolumn{1}{l}{Total duration:} & 30    &       &       &       &       &       &       &  \\
    \multicolumn{1}{l}{ ($\delta$) } & 0.98  &       &       &       &       &       &       &  \\
    \multicolumn{1}{l}{ ($\tilde{\phi}$)} & 1100  &       &       &       &       &       &       &  \\
    \multicolumn{1}{l}{  ($\phi^{Au}$) } & 270   &       &       &       &       &       &       &  \\
    \multicolumn{1}{l}{ ($\phi^{U}$) } & 340   &       &       &       &       &       &       &  \\
    \multicolumn{1}{l}{ ($\phi^{Fe}$) } & 564000 &       &       &       &       &       &       &  \\
    \multicolumn{1}{l}{ ($\phi^{Cu}$) } & 6050000 &       &       &       &       &       &       &  \\
    \multicolumn{1}{l}{ $(\gamma_1,\gamma_2)$} & $(7,36)$     &       &       &       &       &       &       &  \\
    \multicolumn{1}{l}{($\mu^{Fl}$)} & 0.11  &       &       &       &       &       &       &  \\
    \multicolumn{1}{l}{ ($\mu^{U}$)} & 0.79  &       &       &       &       &       &       &  \\
    \multicolumn{1}{l}{ $(\mu^{Cu}_1, \mu^{Cu}_2)$} & $(2.5 ,0)$  &       &       &       &       &       &       &  \\
    \multicolumn{1}{l}{ ( $R^{Fl}_p$)} & 500   &       &       &       &       &       &       &  \\
    \multicolumn{1}{l}{  (${Cu}_p$)} & 0.9   &       &       &       &       &       &       &  \\
    \midrule
    \multicolumn{1}{l}{Ore shipping } & Tonage of ore & \multicolumn{1}{l}{Cu grade} & \multicolumn{1}{l}{Ag grade} & \multicolumn{1}{l}{Fe grade} & \multicolumn{1}{l}{Au grade} & \multicolumn{1}{l}{U grade} & \multicolumn{1}{l}{Fl grade} & \multicolumn{1}{l}{S grade} \\
    1     & 480000 & 1.08  & 1.33  & 14.08 & 1.56  & 32.73 & 1263  & 0.21 \\
    2     & 220000 & 1.86  & 3.81  & 25.9  & 0.47  & 70.69 & 2568  & 0.8 \\
    3     & 970000 & 1.79  & 3.41  & 28.97 & 0.5   & 127.83 & 4500  & 0.74 \\
    4     & 400000 & 0.96  & 2.49  & 25    & 0.49  & 400   & 7500  & 0.8 \\
    5     & 3550000 & 1.37  & 2.02  & 14.21 & 0.31  & 44    & 3000  & 0.5 \\
    6     & 1130500 & 0.93  & 2.13  & 23.76 & 1.25  & 26.73 & 1560  & 0.26 \\
    7     & 5377000 & 1.61  & 2.22  & 16.5  & 0.61  & 31    & 2780  & 0.15 \\
    \midrule
    \multicolumn{1}{l}{Customer requirements} &     ($K_p$)   &       &       &       &       &       &       &  \\
    Parcel 1     & 750000 &       &       &       &       &       &       &  \\
    Parcel 2     & 600000 &       &       &       &       &       &       &  \\
    Parcel 3     & 420000 &       &       &       &       &       &       &  \\
    \bottomrule
    \end{tabular}%
  \label{tab:ins1}}}%
\end{table*}%

\begin{table*}[htbp]
  \centering
  \caption{Parameters of Instance 2}
  \scriptsize
  \scalebox{0.8}{
 \makebox[\linewidth][c]{
    \begin{tabular}{lrrrrrrrr}
    \toprule
    \multicolumn{1}{l}{Number of parcels:} & 3     &       &       &       &       &       &       &  \\
    \multicolumn{1}{l}{Number of stockpiles:} & 7     &       &       &       &       &       &       &  \\
    \multicolumn{1}{l}{Total duration:} & 28    &       &       &       &       &       &       &  \\
    \multicolumn{1}{l}{ ($\delta$) } & 0.98  &       &       &       &       &       &       &  \\
    \multicolumn{1}{l}{ ($\tilde{\phi}$)} & 1100  &       &       &       &       &       &       &  \\
    \multicolumn{1}{l}{  ($\phi^{Au}$) } & 270   &       &       &       &       &       &       &  \\
    \multicolumn{1}{l}{ ($\phi^{U}$) } & 340   &       &       &       &       &       &       &  \\
    \multicolumn{1}{l}{ ($\phi^{Fe}$) } & 564000 &       &       &       &       &       &       &  \\
    \multicolumn{1}{l}{ ($\phi^{Cu}$) } & 6050000 &       &       &       &       &       &       &  \\
    \multicolumn{1}{l}{ $(\gamma_1,\gamma_2)$} & $(7,36)$     &       &       &       &       &       &       &  \\
    \multicolumn{1}{l}{($\mu^{Fl}$)} & 0.11  &       &       &       &       &       &       &  \\
    \multicolumn{1}{l}{ ($\mu^{U}$)} & 0.79  &       &       &       &       &       &       &  \\
    \multicolumn{1}{l}{ $(\mu^{Cu}_1, \mu^{Cu}_2)$} & $(2.5 ,0)$  &       &       &       &       &       &       &  \\
    \multicolumn{1}{l}{ ( $R^{Fl}_p$)} & 400   &       &       &       &       &       &       &  \\
    \multicolumn{1}{l}{  (${Cu}_p$)} & 1   &       &       &       &       &       &       &  \\
    \midrule
    \multicolumn{1}{l}{Ore shipping } & Tonage of ore & \multicolumn{1}{l}{Cu grade} & \multicolumn{1}{l}{Ag grade} & \multicolumn{1}{l}{Fe grade} & \multicolumn{1}{l}{Au grade} & \multicolumn{1}{l}{U grade} & \multicolumn{1}{l}{Fl grade} & \multicolumn{1}{l}{S grade} \\
    1     & 480000 & 1.78  & 1.33  & 14.08 & 1.56  & 32.73 & 1263  & 0.21 \\
    2     & 220000 & 1.86  & 3.81  & 25.9  & 0.47  & 70.69 & 2568  & 0.8 \\
    3     & 970000 & 1.79  & 3.41  & 28.97 & 0.5   & 127.83 & 4500  & 0.74 \\
    4     & 400000 & 1.16  & 2.49  & 25    & 0.49  & 400   & 7500  & 0.8 \\
    5     & 3550000 & 0.77 & 2.02  & 14.21 & 0.31  & 44    & 3000  & 0.5 \\
    6     & 1130500 & 1.23  & 2.13  & 23.76 & 1.25  & 26.73 & 1560  & 0.26 \\
    7     & 53770 & 1.81  & 2.22  & 16.5  & 0.61  & 31    & 2780  & 0.15 \\
    \midrule
    \multicolumn{1}{l}{Customer requirements} &     ($K_p$)   &       &       &       &       &       &       &  \\
    Parcel 1     & 300000 &       &       &       &       &       &       &  \\
    Parcel 2     & 460000 &       &       &       &       &       &       &  \\
    Parcel 3     & 330000 &       &       &       &       &       &       &  \\
    \bottomrule
    \end{tabular}%
  \label{tab:ins2}}}%
\end{table*}%

\begin{table*}[htbp]
  \centering
  \caption{Parameters of Instance 3}
  \scriptsize
  \scalebox{0.8}{
 \makebox[\linewidth][c]{
    \begin{tabular}{lrrrrrrrr}
    \toprule
    \multicolumn{1}{l}{Number of parcels:} & 4    &       &       &       &       &       &       &  \\
    \multicolumn{1}{l}{Number of stockpiles:} & 7     &       &       &       &       &       &       &  \\
    \multicolumn{1}{l}{Total duration:} & 31    &       &       &       &       &       &       &  \\
    \multicolumn{1}{l}{ ($\delta$) } & 0.98  &       &       &       &       &       &       &  \\
    \multicolumn{1}{l}{ ($\tilde{\phi}$)} & 1100  &       &       &       &       &       &       &  \\
    \multicolumn{1}{l}{  ($\phi^{Au}$) } & 270   &       &       &       &       &       &       &  \\
    \multicolumn{1}{l}{ ($\phi^{U}$) } & 340   &       &       &       &       &       &       &  \\
    \multicolumn{1}{l}{ ($\phi^{Fe}$) } & 564000 &       &       &       &       &       &       &  \\
    \multicolumn{1}{l}{ ($\phi^{Cu}$) } & 6050000 &       &       &       &       &       &       &  \\
    \multicolumn{1}{l}{ $(\gamma_1,\gamma_2)$} & $(7,36)$     &       &       &       &       &       &       &  \\
    \multicolumn{1}{l}{($\mu^{Fl}$)} & 0.11  &       &       &       &       &       &       &  \\
    \multicolumn{1}{l}{ ($\mu^{U}$)} & 0.79  &       &       &       &       &       &       &  \\
    \multicolumn{1}{l}{ $(\mu^{Cu}_1, \mu^{Cu}_2)$} & $(2.5 ,0)$  &       &       &       &       &       &       &  \\
    \multicolumn{1}{l}{ ( $R^{Fl}_p$)} & 400   &       &       &       &       &       &       &  \\
    \multicolumn{1}{l}{  (${Cu}_p$)} & 1   &       &       &       &       &       &       &  \\
    \midrule
    \multicolumn{1}{l}{Ore shipping } & Tonage of ore & \multicolumn{1}{l}{Cu grade} & \multicolumn{1}{l}{Ag grade} & \multicolumn{1}{l}{Fe grade} & \multicolumn{1}{l}{Au grade} & \multicolumn{1}{l}{U grade} & \multicolumn{1}{l}{Fl grade} & \multicolumn{1}{l}{S grade} \\
    1     & 5000000	& 1.58	& 1.33	& 14.08	& 1.56&	32.73& 1263 &	0.21 \\
    2     & 4200000	& 1.86	& 3.81 &25.9 & 0.47 & 70.69 & 2568 & 0.8\\
    3     & 9700000	& 1.79	& 3.41 & 28.97	& 0.5 & 127.83 & 4500 & 0.74\\
    4     & 4000000	& 1.16 & 2.49 &	25 & 0.49 & 400	& 7500 & 0.8 \\
    5     & 3550000	& 1.37	& 2.02	& 14.21	& 0.31 &	44	& 3000	& 0.5 \\
    6     & 1130500	& 1.13	& 2.13 & 	23.76& 	1.25& 	26.73&	1560	&0.26\\
    7     & 5377000	& 1.91	& 2.22	& 16.5	& 0.61	& 31 & 	2780 &	0.15\\
    \midrule
    \multicolumn{1}{l}{Customer requirements} &     ($K_p$)   &       &       &       &       &       &       &  \\
    Parcel 1     & 137000 &       &       &       &       &       &       &  \\
    Parcel 2     & 94000 &       &       &       &       &       &       &  \\
    Parcel 3     & 92000 &       &       &       &       &       &       &  \\
    Parcel 4     & 111000 &       &       &       &       &       &       &  \\
    \bottomrule
    \end{tabular}%
  \label{tab:ins3}}}%
\end{table*}%

\end{document}